\documentclass{article}
\usepackage{spconf,amsmath,graphicx}
\usepackage{amssymb}
\usepackage{multirow}
\usepackage{booktabs}
\usepackage[colorlinks,citecolor=blue,linkcolor=blue]{hyperref}
\usepackage{arydshln}
\usepackage[table,xcdraw]{xcolor} 
\usepackage{colortbl}

\title{D$^2$-VR: Degradation-Robust and Distilled Video Restoration with Synergistic Optimization Strategy}
\name{Jianfeng Liang, Shaocheng Shen, Botao Xu, Qiang Hu, Xiaoyun Zhang}
\address{Shanghai Jiaotong University}

\begin{document}
%
\maketitle
\begin{abstract}

The integration of diffusion priors with temporal alignment has emerged as a transformative paradigm for video restoration, delivering fantastic  perceptual quality, yet the practical deployment of such frameworks is severely constrained by prohibitive inference latency and temporal instability when confronted with complex real-world degradations. To address these limitations, we propose \textbf{D$^2$-VR}, a single-image diffusion-based video-restoration framework with low-step inference. To obtain precise temporal guidance under severe degradation, we first design a Degradation-Robust Flow Alignment (DRFA) module that leverages confidence-aware attention to filter unreliable motion cues. We then incorporate an adversarial distillation paradigm to compress the diffusion sampling trajectory into a rapid few-step regime. Finally, a synergistic optimization strategy is devised to harmonize perceptual quality with rigorous temporal consistency. Extensive experiments demonstrate that D$^2$-VR achieves state-of-the-art performance while accelerating the sampling process by \textbf{12$\times$}.

\end{abstract}
\section{Introduction}
\label{sec:intro}

Video restoration has become a cornerstone technology in modern digital media, with applications ranging from smartphone photography to high-definition live streaming. Its primary objective is to recover textures and details from low-quality (LQ) input videos to reconstruct high-quality (HQ) outputs. While traditional restoration methods~\cite{chan2021basicvsr} have established a solid foundation, pre-trained diffusion models serving as powerful generative priors, have demonstrated an exceptional ability to restore textures and semantic details that were previously irrecoverable.
Recently, Diffusion-based video models~\cite{yang2024cogvideox} leverage the powerful representation capability of Diffusion Transformers (DiT) and temporal modeling, gradually evolving into a new paradigm for video restoration. However, these models~\cite{zhou2024upscale, xie2025star} typically incur massive parameter footprints and substantial memory requisites, rendering their deployment on consumer-grade GPUs prohibitive.

Consequently, adapting single-image diffusion frameworks emerges as a resource-efficient alternative~\cite{wang2024exploiting}, yet their frame-wise application inevitably introduces temporal inconsistency. Existing well-established methods incorporate optical flow modules to alleviate flickering artifacts~\cite{rota2024enhancing}. But those frameworks heavily rely on optical flow estimation, which can be severely disrupted by degradation, resulting in unsatisfactory perceptual quality and temporal consistency when processing degraded videos.

Moreover, a critical bottleneck of diffusion models is the requirement for multi-step iterative sampling. While various studies have investigated inference acceleration for diffusion models.
Fast samplers~\cite{lu2022dpm} leverage higher-order numerical solvers to stably compress the inference trajectory to fewer steps. 
Progressive distillation~\cite{salimans2022progressive} accelerates sampling by iteratively mapping a multi-step diffusion process into a condensed few-step student model. Similarly, guidance distillation~\cite{meng2023distillation} distills a computationally intensive condition-guided process into a unified unconditional model, thereby expediting inference while retaining the benefits of guidance. 
Nevertheless, these few-step methods always suffer from over-smoothed details in the generated results.

To address these challenges, we propose \textbf{D$^2$-VR}, a degradation-robust and distilled video restoration framework built upon a single-image diffusion model, which leverages motion-compensated features from previous frames as explicit conditional inputs to guide the restoration of the current frame. As described earlier, standard optical flow estimation is often fragile under severe degradation, we introduce the Degradation-Robust Flow Alignment (DRFA) module. By integrating a confidence-aware attention mechanism, this module adaptively filters unreliable motion cues, ensuring that the conditional input remains accurate even under severe occlusions or noise.

To reconcile the trade-off between fast sampling and temporal stability, we devise a Synergistic Optimization Strategy. Specifically, a feature-based spatial adversarial objective recovers high-frequency textures by leveraging a pre-trained UNet encoder, while a Temporal-LPIPS (T-LPIPS) loss enforces consistent transition dynamics, effectively counterbalancing the flickering artifacts often induced by accelerated generation.

We summarize our primary contributions as follows:
\begin{itemize}
    \item We present a novel framework that adapts arbitrary single-image diffusion priors to video restoration tasks. Integrating degradation-robust motion estimation with tailored optimization, it achieves rigorous temporal consistency with only \textbf{4} inference steps
    \item We introduce a flow alignment module integrated with a confidence-aware attention. By adaptively suppressing unreliable motion cues induced by severe degradations, it significantly enhances both feature alignment accuracy and inter-frame consistency.
    \item We devise a synergistic optimization strategy coupling feature-based spatial adversarial learning with Temporal-LPIPS. This strategy reconciles texture fidelity with temporal stability, effectively eliminating flickering artifacts induced by accelerated generation.
\end{itemize}

\section{Method}
\label{sec:method}

\subsection{Overview}
\begin{figure*}[t]
    \centering
    \includegraphics[width=1.0\linewidth]{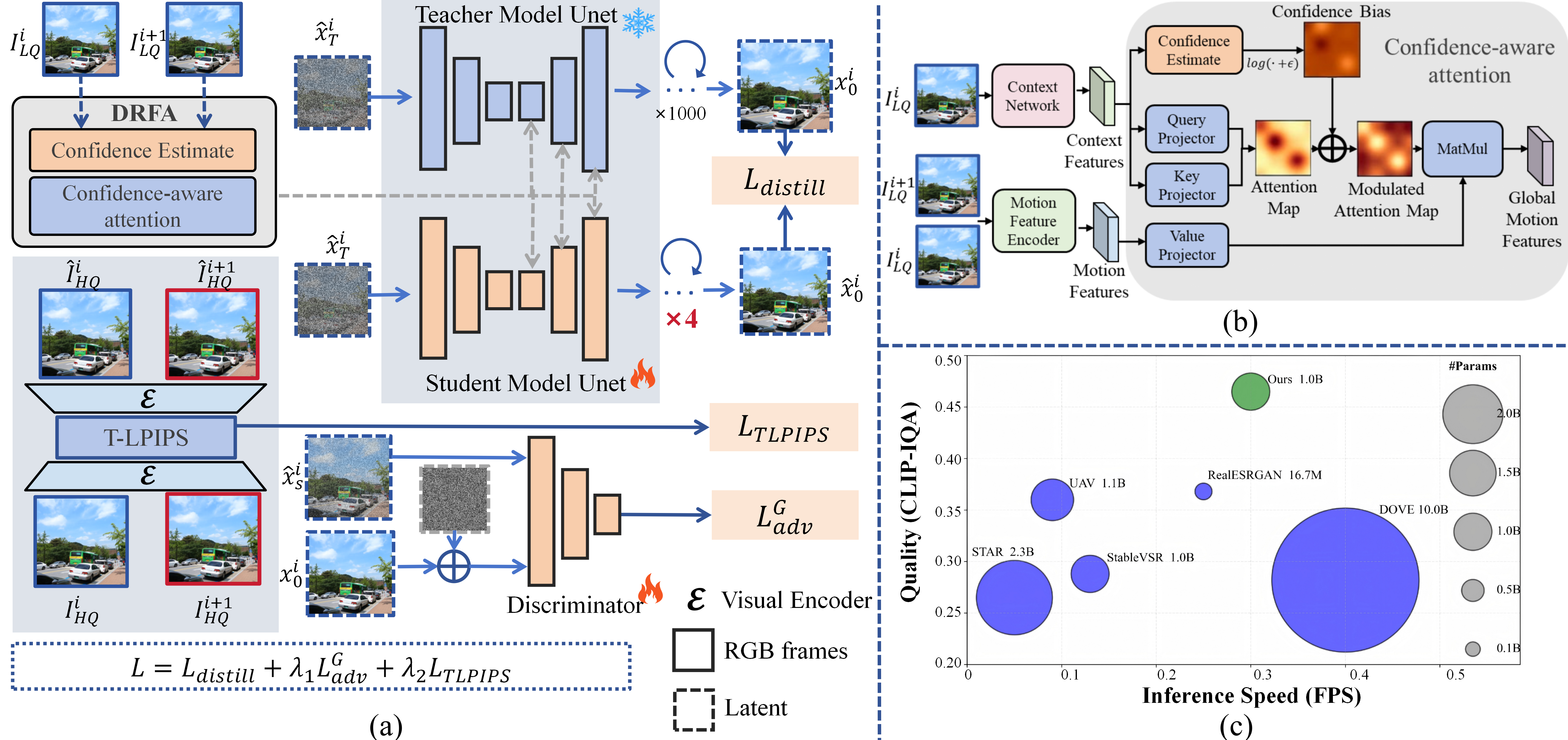}
    \caption{ \textbf{D$^2$VR Overview.} (a) Training and inference pipeline with the computation of the three loss terms. (b) DRFA module architecture. (c) Perceptual quality–speed trade-off comparison between our method and existing approaches, where the x-axis represents inference speed (FPS) and the y-axis represents perceptual quality (CLIP-IQA).}
    \label{fig:1}
\vspace{-3.0mm}
\end{figure*}

We present D$^2$-VR, a degradation-robust and distilled framework equipped with synergistic optimization strategy. Given a low-quality video sequence \{$\mathbf{I}_{LQ}^1$, $\mathbf{I}_{LQ}^2$, $\cdots$, $\mathbf{I}_{LQ}^T$ \} ($\mathbf{I}_{LQ}^t \in \mathbb{R}^{ C \times H \times W}$) containing degradations, our goal is to reconstruct a high-fidelity sequence \{$\mathbf{I}_{HQ}^1$, $\mathbf{I}_{HQ}^2$, $\cdots$, $\mathbf{I}_{HQ}^T$\} that preserves both semantic details and temporal coherence. An overview of the proposed architecture is illustrated in Fig.~\ref{fig:1}(a). Inspired by StableVSR~\cite{rota2024enhancing}, we leverage optical flow-based motion compensation as a  temporal constraint to enforce inter-frame consistency. To mitigate flow instability under degradations, we introduce the Degradation-Robust Flow Alignment (DRFA) module. Leveraging a Confidence-aware Attention Mechanism, it rectifies unreliable motion cues to ensure precise temporal guidance.
We employ an Adversarial Distillation strategy to accelerate the sampling process. By distilling generative priors from a teacher model, the student model retains robust restoration capabilities even within a few-step inference regime, while the integration of an adversarial discriminator further enhances perceptual fidelity. Furthermore, to eliminate inter-frame flickering artifacts exacerbated by such rapid sampling, we incorporate a synergistic optimization strategy.

\subsection{Degradation-Robust Flow Alignment}

A critical challenge for our framework is that conventional optical flow estimation methods often falter in degraded environments. While the Global Motion Aggregation (GMA)~\cite{jiang2021learning} framework attempts to mitigate local inaccuracies via a global attention mechanism, it remains vulnerable under severe degradations. Specifically, the uncontrolled global propagation of noise from corrupted pixels can induce erroneous artifacts within the predicted flow field, ultimately compromising the perceptual quality of the restored video.

To address these issues, we propose the Degradation-Robust Flow Alignment (DRFA) module, which incorporates a confidence-aware attention mechanism to further enhance model robustness against environmental degradations. We introduce a lightweight confidence estimation function $\Phi$ that evaluates the reliability of each spatial location $(i, j)$ based on the context features $f_{ctx}$. The resulting confidence map $C \in \mathbb{R}^{H \times W}$ is defined as:
\begin{equation}
    C = \sigma(\Phi(f_{ctx}))
\end{equation}

where $\sigma(\cdot)$ denotes the Sigmoid activation function, mapping the output to the range $(0, 1)$. Specifically, to suppress the error propagation originating from corrupted local features in degraded scenarios, we augment the vanilla GMA with a confidence-aware gating mechanism. The modulated attention score $S_{i,j}$ is formulated as:
\begin{equation}
    S_{i,j} = \frac{Q_i K_j^T}{\sqrt{d}} + \text{Pos}(i, j) + M_j
\end{equation}

where $M_j = \log(C_j + \epsilon)$ represents the confidence-based bias at position $j$. The introduced bias term $M_j$ acts as a soft-thresholding filter that selectively inhibits information flow from low-fidelity regions. By introducing this quality-aware negative bias prior to attention pooling, our module facilitates the proactive exclusion of contaminated features. Consequently, the synthesized global motion feature $P_i$ at the current location becomes strongly coupled with regions that exhibit both high spatial correlation and superior data confidence. Unlike the original GMA, which performs unconditional aggregation, our approach ensures feature integrity by prioritizing reliable motion cues during the global consensus process.

\subsection{Efficient Adversarial Distillation}

Standard diffusion models necessitate iterative sampling to progressively denoise latent variables. This computationally intensive process incurs prohibitive latency, thereby strictly limiting their applicability in real-world deployment scenarios. To circumvent this, we adopt the Adversarial Distillation paradigm, with the primary objective of compressing the inference trajectory into a rapid few-step regime while preserving high-fidelity generative capabilities. As illustrated in Fig.~\ref{fig:1}(a), a frozen teacher diffusion model serves as a robust source of gradient supervision. We leverage the Score Distillation Sampling (SDS) objective to inherit generative priors by aligning the student to the teacher’s probability distribution, ensuring superior synthesis quality in few-steps settings. Building upon this distillation baseline, we introduce two complementary constraints to resolve the specific challenges of texture over-smoothing and temporal instability.

\textbf{Feature-Based Spatial Adversarial Objective.} While distillation ensures semantic correctness, drastically reducing inference steps often results in over-smoothed textures. To recover high-frequency details, we introduce a feature-based adversarial objective. During training, the generated samples $\hat{\mathbf{x}}_{s}$ and ground-truth latents $\mathbf{x}_{0}$ are passed to the discriminator which aims to distinguish between them. To enable few-step inference, the student model is trained to ensure that its samples at a selected set of timesteps $\mathbf{T}_{student}$ are indistinguishable from real data that has been forward-diffused to the corresponding noise levels. Specifically, we leverage the encoder of a pre-trained UNet as the discriminator. During training, the generator and discriminator are optimized in an alternating adversarial manner

For the generator, the optimization objective is defined as:
\begin{equation}
    \mathcal{L}_{adv}^G = \mathbb{E}_{\mathbf{x},\mathbf{s} \in \mathbb{T}_{student}}[D(\hat{\mathbf{x}}_{s})]
\end{equation}

For the discriminator, the optimization objective is defined as:
\begin{equation}
    \mathcal{L}_{adv}^D = \mathbb{E}_{\mathbf{x},\mathbf{s} \in \mathbb{T}_{student}}[(1-D(\hat{\mathbf{x}}_{s}))+D(\alpha_s \mathbf{x}_{0}+\theta_s\epsilon))]
\end{equation}

where $\mathbf{x}_{0}$ represents the ground truth latents, $\hat{\mathbf{x}}_{s}$ denotes the student's prediction, $\alpha_s$ and $\theta_s$ are hyperparameters of the  diffusion forward process. 

The adversarial objective enables us to fully leverage the adversarial training paradigm, guiding the student model to synthesize sharp high-frequency textures that align with the real data distribution.

\begin{table*}[t]
\vspace{-3.0mm}
\setlength{\tabcolsep}{3pt}
\caption{ Quantitative comparisons on benchmarks, including synthetic (REDS30~\cite{nah2019ntire}), real-world (VideoLQ~\cite{chan2022investigating}). The best and second performances are marked in \textcolor{red}{red} and \textcolor{blue}{blue}, respectively.}
\centering
\resizebox{0.95\textwidth}{!}{\renewcommand{\arraystretch}{1.0}
\begin{tabular}{c|c|ccccccc}
\toprule[1pt]
\textbf{Datasets}           & \textbf{Metrics}  & \textbf{SD $\times$4 Upscaler}  & \textbf{STAR} & \textbf{Real-ESRGAN} & \textbf{UAV} & \textbf{DOVE} & \textbf{StableVSR}  & \textbf{D$^2$-VR(ours)} \\ \hline
\multirow{10}{*}{REDS30}    & PSNR ↑                                & 24.31                             & 22.33    & 23.20          &   23.26           & \textcolor{blue}{24.72}              & 23.19                       & \textcolor{red}{24.72}         \\
                            & SSIM ↑           & 0.621                 & 0.590              & 0.623                     &   0.587          & \textcolor{red}{0.676}              & 0.559                       & \textcolor{blue}{0.653}         \\
                            & LPIPS ↓          & 0.325                     & 0.297              & 0.289                 &   0.315          & \textcolor{blue}{0.289}              & 0.391                       & \textcolor{red}{0.242}         \\
                            & DISTS ↓          & 0.134                 & 0.138              & \textcolor{blue}{0.130}                     &   0.148           & 0.143              & 0.194                       & \textcolor{red}{0.119}         \\ \cdashline{2-9}
                            & MUSIQ ↑          & 51.06               & 59.76              & \textcolor{blue}{64.12}                     &   57.74           & 50.77              & 47.95                       & \textcolor{red}{65.67}         \\
                            & MANIQA ↑         & 0.237               & 0.269              & \textcolor{red}{0.329}                     &   0.276           & 0.229              & 0.241                       & \textcolor{blue}{0.321}        \\
                            & CLIP-IQA ↑       & \textcolor{blue}{0.345}               & 0.265              & 0.368                     &   0.344           & 0.282               & 0.288                       & \textcolor{red}{0.465}         \\
                            & NIQE ↓           & 4.53               & 3.45              & 3.24                     &   \textcolor{blue}{2.85}           & 3.51              & 2.99                       & \textcolor{red}{2.84}               \\ \cdashline{2-9}
                            & tLPIPS ↓         & 86.97               & 13.99              & 25.51                     &   23.33           & \textcolor{red}{12.07}              & 79.03                       & \textcolor{blue}{13.58}         \\
                            & tOF ↓            & 14.39               & \textcolor{blue}{6.680}              & 6.818                     &   28.06           & 7.900              & 11.72                       & \textcolor{red}{4.715}         \\
                            \hline
\multirow{4}{*}{VideoLQ}    & MUSIQ↑          & 31.31                 & 37.72              & 52.98                     & \textcolor{blue}{54.71}             & 48.72              & 31.85                       & \textcolor{red}{58.48}              \\
                            & MANIQA ↑         & 0.149                 & 0.195              & 0.262                     & \textcolor{blue}{0.274}             & 0.240              & 0.147                       & \textcolor{red}{0.278}            \\
                            & CLIP-IQA↑       & 0.173               & 0.238              & 0.292                     & \textcolor{blue}{0.343}             & 0.276              & 0.184                       & \textcolor{red}{0.403}              \\
                            & NIQE↓           & 6.48                     & 5.62              & 4.96                 & \textcolor{red}{3.99}             & 5.02              & 5.34                       & \textcolor{blue}{4.14}             
                            \\
                            \bottomrule[1pt]
\end{tabular}
}
\label{tab:1}
\vspace{-2.0mm}
\end{table*}

\begin{table}[h]
\vspace{-3.0mm}
\setlength{\tabcolsep}{3pt}
\caption{Ablation experiments for DRFA module, quantitative results. Best results in bold text.}
\centering
\resizebox{0.45\textwidth}{!}{
\begin{tabular}{l|lllllllll}
\toprule[1pt]
\textbf{Model} & \textbf{PSNR ↑} & \textbf{SSIM ↑} &  \textbf{NIQE ↓} &  \cellcolor[HTML]{F2F3F5}\textbf{tLPIPS ↓} & \cellcolor[HTML]{F2F3F5}\textbf{tOF ↓}  \\ \hline
RAFT           &  24.57              &   0.643           &  2.89   &     14.44            &    5.118                                     \\
FF             &  24.44              &  0.638         & 3.08        &   19.07               &  5.454                       \\
GMA            &   24.30    &   0.635   &    3.11           &   19.13    &    5.300     \\
\cdashline{1-6}
Ours           &    \textbf{24.72}            &     \textbf{0.653}   &  \textbf{2.84}     &       \textbf{13.58} &      \textbf{4.715}                             \\                                     
\bottomrule[1pt]
\end{tabular}
}
\vspace{-2.0mm}
\label{tab:2}
\end{table}

\begin{table*}[t]
\setlength{\tabcolsep}{3pt}
\caption{Ablation experiments for efficienct adversarial distillation, quantitative results. The best and second performances are marked in \textcolor{red}{red} and \textcolor{blue}{blue}, respectively.}
\centering
\resizebox{0.95\textwidth}{!}{
\begin{tabular}{l|cccccccccc}
\toprule[1pt]
\textbf{Model} & \textbf{PSNR↑} & \textbf{SSIM↑} & \textbf{LPIPS↓} & \textbf{DISTS↓} & \textbf{MUSIQ↑} & \textbf{MANIQA↑} & \textbf{CLIP-IQA↑} & \textbf{NIQE↓} & \cellcolor[HTML]{F2F3F5}\textbf{tLPIPS↓} & \cellcolor[HTML]{F2F3F5}\textbf{tOF↓} \\ \hline
D$^2$-VR w/o $L_{tlpips}$ $L_{adv}$            & \textcolor{red}{26.19}                & \textcolor{blue}{0.698}                & 0.307                & 0.160                  & 49.77                & 0.211                 &  0.218                    & 4.82               & \textcolor{red}{12.20}                                         & \textcolor{red}{4.110}                                      \\
D$^2$-VR w/o $L_{tlpips}$            & 24.05               & 0.611               & \textcolor{blue}{0.251}                & \textcolor{red}{0.114}                & \textcolor{blue}{65.41}                & \textcolor{red}{0.322}                 & \textcolor{red}{0.490}                   & \textcolor{blue}{3.41}               & 68.61                & 4.750                 \\
D$^2$-VR w/o $L_{adv}$         & \textcolor{blue}{26.15}               & \textcolor{red}{0.699}               & 0.296                & 0.155                & 47.26                & 0.189                 & 0.209                   & 4.97               & 13.63                & \textcolor{blue}{4.215}                \\
\cdashline{1-11}
D$^2$-VR(ours)         & 24.72               & 0.653               & \textcolor{red}{0.242}                & \textcolor{blue}{0.119}                & \textcolor{red}{65.67}                & \textcolor{blue}{0.321}                 & \textcolor{blue}{0.465}                   & \textcolor{red}{2.84}               & \textcolor{blue}{13.58}                & 4.715                 \\  
\bottomrule[1pt]
\end{tabular}
}
\label{tab:3}
\end{table*}
\textbf{Synergistic Optimization Strategy.} 
Since the discriminator optimizes perceptual quality without explicit inter-frame reasoning, it is prone to inducing flickering artifacts and temporally inconsistent texture synthesis. To counterbalance the spatial bias inherent in adversarial training, we devise a synergistic optimization strategy. Specifically, we concurrently couple a spatial adversarial objective and a temporal consistency constraint during the distillation process. We leverage Temporal-LPIPS (T-LPIPS)~\cite{chu2020learning} as an explicit supervisor to enforce temporal coherence. Instead of merely minimizing differences between adjacent frames, our approach aligns the magnitude of perceptual quality changes between a pair of consecutive frames in the generated sequence and the corresponding frame pair in the ground-truth high-quality video. To improve the accuracy of alignment, we compute the perceptual quality changes in the RGB space. The loss is defined as:
\begin{equation}
    \mathcal{L}_{T-LPIPS} = \sum_{t=1}^{T} \| (\phi(\hat{I}_{}^t) - \phi(\hat{I}_{}^{t-1})) - (\phi(I_{HQ}^{t}) - \phi(I_{HQ}^{t-1})) \|_2^2
\end{equation}
where $\phi(\cdot)$ denotes the perceptual feature extractor. Integrating the aforementioned constraints, the total optimization objective for the adversarial distillation framework is defined as follows:
\begin{equation}
\mathcal{L} = \mathcal{L}_{distill}+\lambda_1\mathcal{L}_{adv}^G+\lambda_2\mathcal{L}_{T-LPIPS}
\end{equation}
By harmonizing these constraints, the proposed strategy facilitates the generation of realistic textures without compromising temporal coherence. The synergy among these tailored losses enables our efficient adversarial distillation framework to deliver remarkable restoration quality and temporal stability, all while operating in a rapid few-step setting.

\section{Experiments}
\label{sec:experiment}

\subsection{Implementation details}
D$^2$-VR is built upon Stable Diffusion x4 Upscaler~\cite{rombach2022high}, a single-image super-resolution model based on Stable Diffusion 2.1. During implementation. We first train the DRFA module using a synthetically degraded optical-flow dataset. 
Then we freeze the DRFA module and perform Efficient Adversarial Distillation on the entire network for 5000 steps, where the adversarial loss weight $\lambda_1$ is set to 0.05 and the tLPIPS loss weight $\lambda_2$ is set to 0.1. The Discriminator is the encoder of a pre-trained UNet in Stable Diffusion 2.1. Targeted inference steps are 4 steps. (Targeted set of timesteps is  $\mathbf{T}_{student}=\{750,500,250,0\}$ ) 
Training is conducted on a single NVIDIA A100 GPU using the Adam optimizer, with a batch size of 8 and a fixed learning rate of 2e-5.

\subsection{Datasets and evaluation metrics}
For evaluating our video restoration performance, we conduct experiments on both synthetic~\cite{nah2019ntire} and real-world ~\cite{chan2022investigating} benchmark datasets.

On synthetic dataset with available ground-truth references, we adopt a comprehensive evaluation protocol that includes both full-reference and no-reference metrics. For real-world datasets lacking ground-truth references, only no-reference metrics are utilized. Specifically, PSNR and SSIM are employed to assess pixel-level accuracy; LPIPS~\cite{zhang2018unreasonable}, DISTS~\cite{ding2020image}, MUSIQ~\cite{ke2021musiq}, MANIQA~\cite{yang2022maniqa}, CLIP-IQA~\cite{wang2023exploring}, and NIQE~\cite{mittal2012making} are used to evaluate perceptual quality; tOF~\cite{chu2020learning} and tLPIPS~\cite{chu2020learning} are adopted to measure temporal consistency. 

\subsection{Comparison with state-of-the-art methods}
To evaluate the performance of our framework, we compare D$^2$-VR against state-of-the-art video restoration methods, including Real-ESRGAN~\cite{wang2021real}, SD $\times$4 Upscaler~\cite{rombach2022high}, STAR~\cite{xie2025star}, Upscale-a-Video (UAV)~\cite{zhou2024upscale},
DOVE~\cite{chen2025dove},
StableVSR~\cite{rota2024enhancing}.

\textbf{Quantitative Results.} Tab.~\ref{tab:1} shows that D$^2$-VR achieves superior performance compared to existing state-of-the-art methods across almost all evaluation metrics. D$^2$-VR achieves the best performance on perceptual metrics such as LPIPS, MUSIQ, and CLIP-IQA, as well as on the temporal consistency metric tOF. These results clearly demonstrate that our method achieves an excellent trade-off between perceptual quality and temporal consistency.

\begin{figure}[t]
    \centering
    \includegraphics[width=1.0\linewidth]{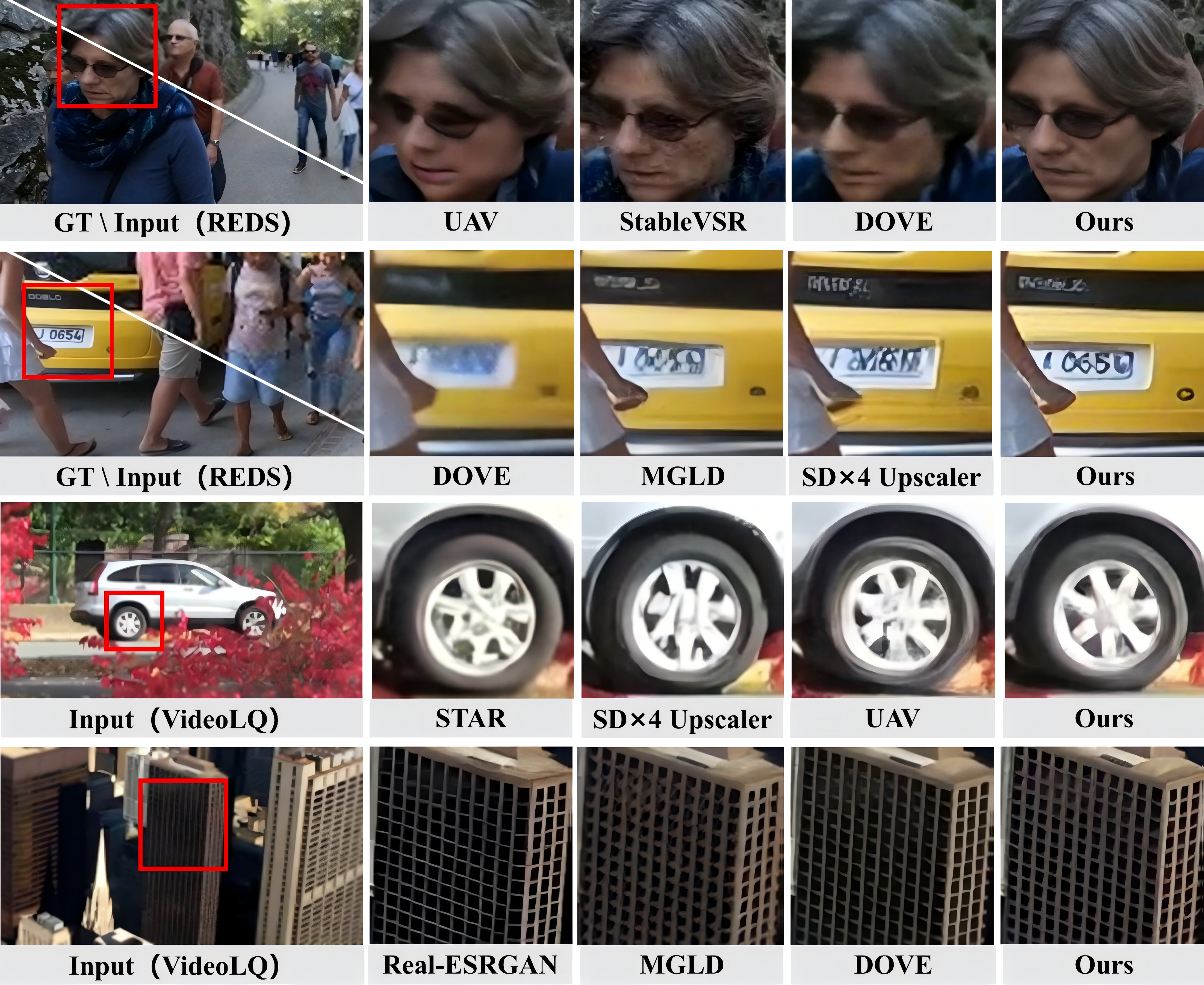}
    \caption{Qualitative comparisons results on benchmarks with existing methods.}
    \label{fig:2}
\vspace{-2.0mm}
\end{figure}

\textbf{Qualitative Results.} Fig.~\ref{fig:2} shows our method generates outputs with richer textures and a capacity for higher fine-detail generation. The restoration frames exhibit high fidelity for complex local structures, with more plausible contours and fewer artifacts. Specifically, the results for faces appear more realistic and clearer, while those for text are more recognizable.

Fig.~\ref{fig:1}(c). illustrates a comprehensive comparison between our method and existing methods in terms of model size, inference speed, and restoration quality.
Methods employing one-step diffusion models, such as DOVE, achieve higher inference speed than D$^2$-VR. However, our method exhibits clear advantages in terms of model lightweightness and restoration quality. 

\subsection{Ablation study}

\textbf{Degradation-Robust Flow Alignment.}
To further examine the contribution of the confidence-aware attention mechanism in DRFA, we conduct an ablation study by replacing DRFA with motion estimation modules based on RAFT~\cite{teed2020raft}, GMA~\cite{jiang2021learning}, and FlowFormer++ (FF)~\cite{shi2023flowformer++}. The results are presented in Tab~\ref{tab:2}. The model with DRFA module achieves the best performance in terms of both fidelity and temporal consistency. The proposed confidence-aware attention mechanism enhances the robustness of optical flow estimation under degraded video conditions, thereby leading to more accurate motion estimation and improved temporal consistency.

\textbf{Synergistic Optimization Strategy.}
During adversarial distillation training, we aim to verify the effectiveness of both the feature-based spatial adversarial objective and the temporal consistency constraint. We compare four variants: removing the adversarial generator loss (D$^2$-VR w/o $\mathcal{L}_{adv}^G$), removing the T-LPIPS loss (D$^2$-VR w/o $\mathcal{L}_{tlpips}$), removing both objectives (D$^2$-VR w/o $\mathcal{L}_{adv}^G$ $\mathcal{L}_{tlpips}$), and ours (D$^2$-VR). The results are presented in Tab~\ref{tab:3}.

For D$^2$-VR w/o $\mathcal{L}_{adv}^G$, MUSIQ drops from 65.67 to 47.26 and MANIQA drops from 0.321 to 0.189, compared with D$^2$-VR.
Removing the adversarial generator loss leads to significant decline in perceptual quality, which can be attributed to over-smoothed restoration results.
In contrast, removing the T-LPIPS loss severely disrupts temporal consistency. For D$^2$-VR w/o $\mathcal{L}_{tlpips}$, tLPIPS increases from 13.58 to 68.61 compared with D$^2$-VR.

Notably, D$^2$-VR does not attain the best results across all metrics. D$^2$-VR w/o $\mathcal{L}_{adv}^G$ $\mathcal{L}_{tlpips}$ achieves the best performance on the temporal consistency metrics tLPIPS and tOF. We speculate that over-smoothed restoration results make inter-frame temporal inconsistencies less noticeable. And the variants without adversarial generator loss achieve the best performance on SSIM and PSNR. This is because the introduction of adversarial loss encourages perceptually plausible high-frequency details, which tend to result in lower pixel-wise fidelity scores.  
These results further confirm that our method achieves a favorable trade-off between temporal consistency and perceptual quality.

\section{Conclusions}
\label{sec:conclusions}

In this paper, we present D$^2$-VR, a degradation-robust and efficiently distilled framework for video restoration built upon single-image diffusion models. To enhance temporal consistency under degradation, we introduce the Degradation-Robust Flow Alignment (DRFA) module, which incorporates a confidence-aware attention mechanism to selectively suppress unreliable motion cues. To overcome the inherent inefficiency of iterative diffusion sampling, we further adopt an adversarial distillation paradigm, compressing the sampling trajectory into a few-step regime while preserving generative capability. Crucially, we propose a synergistic optimization strategy that synergistically combines a feature-based spatial adversarial objective with a temporal consistency constraint which effectively reconciles the trade-off between texture sharpness and temporal consistency.

%


\bibliographystyle{IEEEbib}
\bibliography{strings,refs}

\end{document}